# DAMMI:Daily Activities in a Psychologically Annotated Multi-Modal IoT dataset

Mohsen Falah Rad[1,*], Kamrad Khoshhal Roudposhti[1], Mohammad Hassan Khoobkar[1], Mohsen Shirali[2], Zahra Ahmadi[3], Carlos Fernández-Llatas[4,5]

[1] Department of Computer Engineering, Lahijan Branch, Islamic Azad University, Lahijan, Iran

[2] Department of Computer Science and Engineering, Shahid Beheshti Univeristy

[3] Research Centre for Information Systems Engineering (LIRIS), KU Leuven, Leuven, Belgium

[4] Process Mining 4 Health Lab?SABIEN-ITACA Institute, Universitat Politècnica de València, Valencia, Spain

[5] Department of Clinical Sciences and Intervention and Technology (CLINTEC), Karolinska Institute, Stockholm, Sweden

*Corresponding Author: Mohsen Falah Rad (Email address: mo.falahrad@iau.ac.ir)

## Abstract

The growth in the elderly population and the shift in the age pyramid have increased the demand for healthcare and well-being services. To address this concern, alongside the rising cost of medical care, the concept of "ageing at home" has emerged, driven by recent advances in medical and technological solutions. Smart systems and tools—often leveraging the Internet of Things (IoT)—are proposed to enable continuous and real-time health monitoring, assisting healthcare professionals and medical systems. Experts in computer science, communication technology, and healthcare have collaborated to develop affordable health solutions by employing sensors in living environments, wearable devices, and smartphones, in association with advanced data mining and intelligent systems with learning capabilities, to monitor, analyze, and predict the health status of elderly individuals.

However, implementing intelligent healthcare systems and developing analytical techniques requires testing and evaluating algorithms on real-world data. Despite the need, there is a shortage of publicly available datasets that meet these requirements. To address this gap, we present the DAMMI dataset in this work, designed to support researchers in the field. The dataset includes daily activity data of an elderly individual collected via home-installed sensors, smartphone data, and a wristband over 146 days. It also contains daily psychological reports provided by a team of psychologists. Furthermore, the data collection spans significant events such as the COVID-19 pandemic, New Year's holidays, and the religious month of Ramadan, offering additional opportunities for analysis. In this paper, we outline detailed information about the data collection system, the types of data recorded, and pre-processed event logs. This dataset is intended to assist professionals in IoT and data mining in evaluating and implementing their research ideas.

**Keywords:** Internet of Things, Dataset, elderly care, smart home, Activities of Daily Living.

# 1 Introduction

The percentage of individuals aged 65 or older in 2022 was nearly 10% of the population and it is projected to increase to 16% in 2050 [1]. This demographic shift, driven by advancements in medicine and technology, presents several challenges for developed societies. These challenges include a rise in age-related diseases, further dependence on caregivers, a shortage of trained healthcare professionals, and an increase in healthcare costs [2]. As people age, they are more likely to experience cognitive, mental, and physical health issues. With these difficulties that future healthcare systems have, a failure to promptly identify symptoms of age-related diseases or to continuously monitor treatment can significantly reduce the quality of life for older adults, impacting their health and ability to live independently [3], [4], [5].

Moreover, elderly individuals often prefer to spend their later years in their own homes rather than in nursing homes, as maintaining independence contributes to a better quality of life. However, this preference increases the cost of care-giving. Consequently, the development of assistive technologies in home environments has become essential. One of these technologies that is being developed by experts in the computer and communication fields is "Ambient Assistive Living (AAL)". AAL creates a sensitive, adaptive, and responsive digital environment that supports the elderly in living independently. It integrates information technology, communication systems, and smart technologies into daily life, enabling older adults to remain active and autonomous for longer periods. The global AAL market is expected to reach $13.74 billion by 2027 [6].

By utilizing home-based technologies, like AAL, daily living patterns can be continuously monitored, enabling the detection of abnormalities that may signal the onset of physical or mental health issues, or other unpredictable situations [7]. For example, a fall can be classified as an unexpected event where an elderly person ends up on the ground [8]. In addition, intelligent systems can analyze data to detect and predict mobility, cognitive, and psychological issues. Hence, health-related data is invaluable for assessing and forecasting changes in mobility or memory, aiding in the early diagnosis of conditions such as Alzheimer's [9] and dementia [10]. Moreover, irregularities in sleep patterns, identified through smart home data, can help predict and diagnose conditions such as heart attacks, high blood pressure, diabetes, obesity, stress, and even stroke[11]. Besides, in recent years, many individuals have adopted wearable fitness trackers or pedometers to monitor physical activity (PA), often in pursuit of a goal such as increasing cardiovascular strength, losing weight, or improving overall health [12].

In light of this promising trend, industry experts and academic researchers must prioritize the development of intelligent systems and advanced data mining techniques. These technologies are essential for collecting and analyzing data to provide affordable and efficient healthcare services. They enable personalized healthcare, allowing for treatments and services tailored to an individual's unique physical state, emotional condition, and needs. It is crucial to focus on developing intelligent, smart solutions that can scale with growing needs to prevent healthcare systems from becoming overwhelmed by increasing demands in the future.

However, to ensure these intelligent systems are reliable and effective, they must be thoroughly tested and assessed. It is vital to confirm that they can perform their intended tasks accurately and consistently. These systems should evolve interactively, adapting to the needs of users and experts, and gradually integrating into people's lives while improving their functionality. Evaluating them under realistic conditions is essential to determine how well they can support healthcare professionals and facilitate timely medical interventions. Once these systems have gained the trust of both healthcare experts and the general public, they will be more readily accepted and integrated into everyday life. Therefore the first critical step in developing intelligent healthcare systems—powered by IoT, data mining, and artificial intelligence—must be their rigorous evaluation using real-world data under realistic circumstances.

Access to real-world datasets collected in everyday scenarios is essential for the development of healthcare systems and algorithms, especially those involving IoT devices. These large, diverse datasets provide richer contexts for data mining and analytical techniques, improving the identification of patterns, predicting outcomes, and enhancing diagnostic accuracy and patient monitoring. Additionally, the accuracy of IoT devices in collecting and generating real-time data can be verified by comparing their data streams with those from well-established datasets. This comparison ensures more reliable and responsive healthcare solutions, particularly for dealing with complex conditions and diverse patient populations.

Large datasets also play a key role in improving the robustness of models by making them more resistant to noise, device errors, and anomalies. Healthcare IoT systems often capture imperfect or incomplete data due to device limitations or patient non-compliance. With access to large datasets, algorithms are able to filter out inconsistencies, focusing on meaningful trends and improving the system's overall performance. This is especially critical for real-time analytics, ensuring that healthcare providers can make timely, data-driven decisions to optimize patient care.

However, the lack of rich, publicly available physical datasets has made it challenging to fully assess the performance of IoT technologies—such as smart environments and wearable devices—and to apply advanced data mining techniques. This scarcity has led to limited evaluations, often relying on synthetic data. Therefore, generating and sharing comprehensive datasets is crucial for advancing robust and usable healthcare technologies [13]. By creating and making these datasets publicly available, we foster greater collaboration and accelerate technology evaluation.

In response to this need, we have implemented a comprehensive IoT testbed equipped with ambient sensors, a wristband and a smartphone to monitor Activities of Daily Living (ADLs). This document presents detailed information on this dataset that we are going to share publicly, contributing a valuable resource to the research community.

## 2 Existing IoT and smart home datasets

Creating a comprehensive dataset involves a significant amount of effort, particularly when building physical testbeds and deploying real devices for data collection. In [13], Cook et al. shared valuable insights from their experience in generating smart home datasets, highlighting the many challenges they encountered. The process is not only time-consuming but also requires careful planning, particularly in selecting appropriate sensors and ensuring the data collection is well-planned. It is often recommended that such efforts be carried out in multiple phases to ensure quality and address potential issues as they arise [14].

Due to these challenges, the number of available datasets is still quite limited. Many are either not publicly accessible or involve data collected over short periods or in controlled, laboratory settings. Despite these constraints, several valuable efforts have been made to gather and share datasets that help advance research in this field. These datasets are instrumental for researchers and developers seeking to improve smart healthcare solutions and IoT-based systems.

In Table 1, we provide a list of well-known available datasets that have been frequently used in the literature. The table also details the data collection methods and the types of sensors employed in each case.

# 3   The Detailed description of the dataset

The dataset proposed in this study is collected by an IoT system and contains data from a smart home equipped with ambient sensors, wristband data and smartphone information on the usage of mobile applications. In addition, daily psychological reports are also collected and provided. All of the data modalities are gathered for a period of 146 days in the timeframe between 09/01/2020 and 02/06/2020 from a 60-year-old female subject.

During the collection period, multiple exciting events happened that influenced the behavior of the resident and possibly led to higher variability and complexity in the collected data. There was heavy snow on the 35th day from the beginning of the collection process, due to which the electricity was cut off for several hours. This event also changed the habit of the subject for a few days. The start of the Covid-19 pandemic and a lockdown period from the 40th day in the region affects the habits of the participant. From the 109th day to the 137th day was the Ramadan month; therefore, the resident changed her habits such as meal times, praying time and duration, and sleeping intervals.

In the following more details on the raw data sources and pre-processed extracted event logs are described in detail.

*Table 1 a list of most frequently used IoT datasets for research on AALs in the literature*

| Dataset name | description | Data and collection devices | Access |
|---|---|---|---|
| CASAS smart home datasets [15] | CASAS datasets are collected by Center of Advanced Studies in Adaptive System at Washington State University and include many different smart home setups with different sensor configurations such as PIRs, temperatures, contact switches and pressures. Also, datasets with smartwatch data are available on the website of this project. | PIRs, temperatures, contact switches and pressures | Public |
| Intel Berkeley Research Lab Sensor Data [16] | Intel Berkeley Research Lab Sensor Data. This data is collected from 54 sensors deployed in the Intel Berkeley Research lab between February 28th and April 5th, 2004. | Mica2Dot sensors with weatherboards collected timestamped topology information, along with humidity, temperature, light, and voltage values once every 31 seconds. | Public |
| CART study [17] | The CART study from ORCATECH Life Lab. An array of sensors and smart devices are installed in a home that is enrolled in the CART research study. Data was collected between 2017 - 2020. | This includes data about mobility (walking speed, movement between rooms), socialization (outings, phone calls, emails sent), medication adherence, sleeping patterns and physiologic function (BMI, pulse) | Private |
| 1-Year in-Home Monitoring [18] | 1-Year in-Home Monitoring Technology by Home-Dwelling, Family Caregivers, and Nurses. The system continuously monitored Older Adults' daily activities at home by an ambient sensor system and health-related events by wearable sensors. | Ambient sensor system (DomoCare®) to monitor activities (e.g., mobility, sleep habits, fridge visits, door events) and health-related data (ECG signal, heart rate, HRV, skin temperature, respiration rate, physical activity and mobility) by Fitbit Activity tracker and ECG sensor. | Private |
| eSense smart home testbed [19] | The dataset contains PIR sensor readings from a solo-resident house for two different durations, a 70-days dataset and a 21-days dataset. | PIR sensors for movement detection | Public |
| TIHM [20] | An open dataset for remote healthcare monitoring in dementia. The data was collected from the homes of 56 People Living with Dementia (PLWD) and associated with events and clinical observations (daily activity, physiological monitoring, and labels for health-related conditions). The study recorded an average of 50 days of data per participant, totalling 2803 days. | Sleep, physiological and activity data | Public |
| The proposed dataset | Activities of Daily Living collected by ambient sensors from a solo-resident house with a 60-year-old woman for 147 days. The dataset also includes smartphone application usage and wristband data including sleep-related information, daily steps and active time. In addition, a daily psychological report acquired by a questionnaire is also available for each day which can be used for mood analysis. | Ambient sensors (PIRs and contact switches), information from a wristband, a smartphone and daily psychological reports. | Public |

## 3.1 Data sources

For data collection, we took advantage of multiple sensor devices together to collect data simultaneously. These different data modalities are selected to capture a more comprehensive view of the user and provide us with a multi-perspective view of daily life.

### 3.1.1 Ambient Sensors

These sensors installed in the house are used to perceive the environment and the activities performed by the resident during the day. The installed devices are 15 binary sensors positioned on furnishing elements, appliances, and doors. Indeed, the ambient sensorial information includes readings from multiple PIR sensors showing the presence of the subject in different areas of the solo-resident house, a power usage sensor indicating TV usage, contact sensors highlighting the opening and closing of the Bathroom, WC and closet doors and a gas detection sensor detecting cooking activity.

The types of binary sensors that were used to collect the required data are summarised in Table 2 and their locations are depicted in Figure 1. These sensors sent a value of *one* to the data registering system when any changes in the environment captured. The data storage system was designed so that if the value of one of the sensors changed, all values of the sensors were recorded simultaneously. Figure 2 shows a sample of how the data were stored in the system.

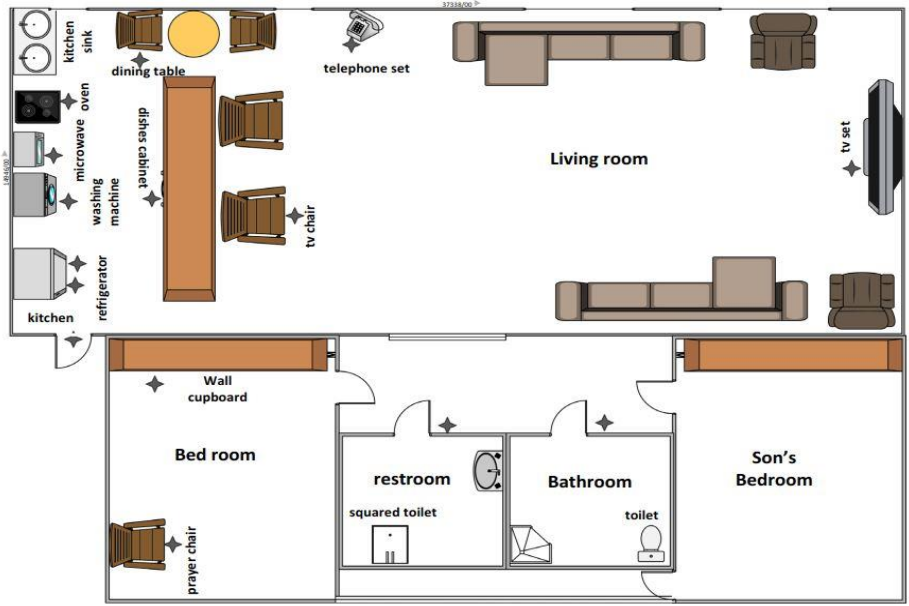

*Figure 1. The floorplan of the house and the placement of ambient sensors*

*Table 2 Installed sensor types and locations in the house*

| Location of the sensor | Sensor type |
|---|---|
| Oven | MQ-9 gas sensor |
| In the refrigerator | IR Sensor |
| In the freezer | IR Sensor |
| In the microwave | IR Sensor |
| Praying chair | IR Sensor |
| Kitchen cabinet | IR Sensor |
| Landline phone | IR Sensor |
| In entry and exit | IR Sensor |
| Washing machine | IR Sensor |
| Dining chair | IR Sensor |
| TV chair | IR Sensor |
| TV | Power usage sensor |
| Bathroom | Micro-switch |
| Restroom | Micro-switch |
| Closet | Push-button |

| Date/Time | TV. | Tell | TV chair | Dining chair | bathroom | Restroom | closet | Exit | Praying chair | refrigerator | freezer | Washing machine | microwave | oven | Kitchen cabinet |
|---|---|---|---|---|---|---|---|---|---|---|---|---|---|---|---|
| 1/8/2020 23:04:33 | 0 | 0 | 0 | 0 | 0 | 0 | 1 | 0 | 0 | 0 | 0 | 0 | 0 | 0 | 0 |
| 1/8/2020 23:05:06 | 0 | 0 | 0 | 0 | 0 | 0 | 0 | 0 | 0 | 0 | 0 | 0 | 0 | 0 | 0 |
| 1/8/2020 23:05:39 | 0 | 0 | 0 | 0 | 1 | 0 | 0 | 0 | 0 | 0 | 0 | 0 | 0 | 0 | 0 |
| 1/8/2020 23:07:21 | 0 | 0 | 0 | 0 | 0 | 0 | 0 | 0 | 0 | 0 | 0 | 0 | 0 | 0 | 0 |
| 1/8/2020 23:07:32 | 0 | 0 | 0 | 0 | 0 | 0 | 0 | 0 | 0 | 0 | 1 | 0 | 0 | 0 | 0 |
| 1/8/2020 23:07:45 | 0 | 0 | 0 | 0 | 0 | 0 | 0 | 0 | 0 | 0 | 0 | 0 | 0 | 0 | 0 |
| 1/8/2020 23:09:20 | 0 | 0 | 0 | 0 | 0 | 0 | 0 | 0 | 0 | 0 | 0 | 0 | 0 | 0 | 0 |
| 1/9/2020 0:49:38 | 1 | 0 | 0 | 0 | 0 | 0 | 0 | 0 | 0 | 0 | 0 | 0 | 0 | 0 | 0 |

*Figure 2 A schematic view of raw sensor readings recorded on the data storage*

### 3.1.2 Wearable sensor data

A Xiaomi Mi Band-3 wristband was used to provide information about daily activity time, walking steps and distance, calories burned and sleep-related data such as night sleep time, duration and quality (deep/light sleep or awake time durations).

### 3.1.3 Mobile data

With the aid of a mobile application, smartphone usage information is collected. The timestamp of mobile usage and the name of the application, which is used, is logged in the raw data.

## 3.2 Data annotations and feature extraction

### 3.2.1 Psychological labelling and daily reports

In addition to the sensorial data, a daily psychological labelling approach is also used by an expert team to determine psychological information for the period of data collection. The team of psychologists proposed a new questionnaire for the psychological-based labelling process, suitable to the experiment scenario and its limitations. This questionnaire is designed according to the well-known standard questionnaires in the literature including

 - The General Health Questionnaire (GHQ-28),

 - Beck Depression Inventory, and

 - PSQI (Pittsburgh Sleep Quality Index)

A psychologist checked the psychological status of the person on a daily basis, and then a team containing three different psychologists answered a predetermined questionnaire with 39 items in seven categories by considering the information from the home visit observation and interaction with the subject as well as the insight provided by the smart facilities. Each question had three options, high (score of three), average (score of two), and low (score of one).

## 4 Preprocessed Event Logs

### 4.1 The location event log

Using ambient sensor data, we have conducted a pre-processing phase to extract the person's presence in different areas of the house. As illustrated in Figure 1, the sensors are installed in six different areas of a house. To discover the exact location of the resident in different areas of the home by using ambient sensorial information, we can look at the locations in which sensors are mounted, and use the timestamp of captured events' to infer the person's location at different times.

For instance, an event captured by the bedroom's closet sensor means that the subject is in the bedroom at that time. Then, by co-relating each sensor to its installation location, every captured event could be mapped to a location. Therefore, if we have the whole sequence of captured events for some time, then, the person's presence in different areas of the house and even the sequence of his/her movements could be inferred.

Thus, the house is partitioned into seven different areas and a set of simple rules are used over the events generated by the PIR and contact sensors to discover the subject's presence in these areas. The timestamp of first captured event and the last event in each area are then considered for determining the events' start and end time and the following labels are assigned to the discovered events at each place; Kitchen, LivingRoom, Bedroom, Corridor, WC, Bathroom and Entrance. Figure 3 presents a part of ambient location event log. This event log includes 12494 records for the whole period of data collection and can be used in studying mobility patterns, behaviour analysis, behaviour anomaly detection and so on.

| Locations | StartTime | EndTime | Duration |
|---|---|---|---|
| Bedroom | 1/9/2020 0:00:00 | 1/9/2020 0:50:20 | 0:50:20 |
| Kitchen | 1/9/2020 0:50:20 | 1/9/2020 1:37:05 | 0:46:45 |
| Bedroom | 1/9/2020 1:37:05 | 1/9/2020 4:45:46 | 3:08:41 |
| Bathroom | 1/9/2020 4:45:46 | 1/9/2020 4:49:16 | 0:03:30 |
| Corridor | 1/9/2020 4:49:16 | 1/9/2020 4:54:03 | 0:04:47 |
| Bedroom | 1/9/2020 4:54:03 | 1/9/2020 8:26:25 | 3:32:22 |
| Bathroom | 1/9/2020 8:26:25 | 1/9/2020 8:28:39 | 0:02:14 |
| Corridor | 1/9/2020 8:28:39 | 1/9/2020 8:33:04 | 0:04:25 |
| Kitchen | 1/9/2020 8:33:04 | 1/9/2020 9:08:22 | 0:35:18 |

*Figure 3. The ambient location event log*

## 4.2 Activity and location event log

In addition to the location events, since some sensors (such as the TV sensor, kitchen appliance sensors, etc.) are just triggered when the subject performs specific activities, hence, besides finding the location of the subject, the sensor readings can also be used to discover the incidence time and duration of those activities. These activities, which are performed by using specific instruments, are known as instrumented Activity Daily Livings (iADL) and are investigated in several studies and projects.

Using the sensor readings, we have detected the discoverable activities for the whole dataset period. These activities are Sleeping, Praying, Sitting, Sitting and Watching TV, Eating, Eating and Watching TV, Cooking, Using Telephone and Chores. The activity and location event log includes 18427 records a part of it is represented in Figure 4. The activity and location event log can be used for research purposes like evaluation of ADL discovery methods, activity pattern recognition and detection of changes in daily routines.

| Location/Events | Date | StartTime | EndTime | Duration |
|---|---|---|---|---|
| Bedroom | 1/9/2020 | 1/9/2020 0:00:00 | 1/9/2020 0:49:20 | 0:49:20 |
| Corridor | 1/9/2020 | 1/9/2020 0:49:20 | 1/9/2020 0:49:25 | 0:00:05 |
| LivingRoom | 1/9/2020 | 1/9/2020 0:49:25 | 1/9/2020 0:50:00 | 0:00:35 |
| Kitchen | 1/9/2020 | 1/9/2020 0:50:00 | 1/9/2020 1:36:59 | 00:46:59 |
| LivingRoom | 1/9/2020 | 1/9/2020 1:37:00 | 1/9/2020 1:37:01 | 00:00:01 |
| Corridor | 1/9/2020 | 1/9/2020 1:37:02 | 1/9/2020 1:37:03 | 00:00:01 |
| Bedroom | 1/9/2020 | 1/9/2020 1:37:04 | 1/9/2020 4:44:59 | 03:07:55 |
| BR-Sleeping | 1/9/2020 | 1/9/2020 1:37:04 | 1/9/2020 4:27:05 | 02:50:01 |
| Corridor | 1/9/2020 | 1/9/2020 4:45:00 | 1/9/2020 4:45:01 | 00:00:01 |
| Bathroom | 1/9/2020 | 1/9/2020 4:45:02 | 1/9/2020 4:48:59 | 00:03:57 |
| Corridor | 1/9/2020 | 1/9/2020 4:49:00 | 1/9/2020 4:53:59 | 00:04:59 |
| Bedroom | 1/9/2020 | 1/9/2020 4:54:00 | 1/9/2020 8:25:59 | 03:31:59 |
| BR-Praying | 1/9/2020 | 1/9/2020 4:54:03 | 1/9/2020 6:05:21 | 01:11:18 |
| BR-Sleeping | 1/9/2020 | 1/9/2020 6:05:21 | 1/9/2020 8:25:59 | 02:20:38 |

*Figure 4. Activity and location event log.*

## 4.3 Wristband event log

The wristband event log contains daily information collected by the wristband including the daily steps, distance, activity time, start sleep time and wake up time for sleeps during night and afternoon, the amount of deep and light sleep. This event log is appropriate for studying the sleep habits or daily activity level.

## 4.4 Mobile app usage event log

The information about used mobile applications during the data collection period are provided in this log. This log includes 12670 records showing which type of mobile applications have been used and for how long. This event log is useful for detecting mobile usage patterns and studying mobile usage habits.

| App Type | Date | StartTime | EndTime | Duration |
|---|---|---|---|---|
| Tools | 1/9/2020 | 1/9/2020 4:27:05 | 1/9/2020 4:27:28 | 0:00:23 |
| Religious | 1/9/2020 | 1/9/2020 5:53:04 | 1/9/2020 5:53:26 | 0:00:22 |
| Social Media | 1/9/2020 | 1/9/2020 8:49:12 | 1/9/2020 8:51:17 | 0:02:05 |
| Social Media | 1/9/2020 | 1/9/2020 9:55:28 | 1/9/2020 9:55:29 | 0:00:01 |
| Tools | 1/9/2020 | 1/9/2020 9:55:32 | 1/9/2020 9:56:08 | 0:00:36 |
| Tools | 1/9/2020 | 1/9/2020 10:54:43 | 1/9/2020 10:54:47 | 0:00:04 |
| Call | 1/9/2020 | 1/9/2020 10:54:50 | 1/9/2020 10:54:58 | 0:00:08 |
| Call | 1/9/2020 | 1/9/2020 10:54:58 | 1/9/2020 10:54:59 | 0:00:01 |
| Call | 1/9/2020 | 1/9/2020 10:55:46 | 1/9/2020 10:56:11 | 0:00:25 |
| Management | 1/9/2020 | 1/9/2020 11:11:56 | 1/9/2020 11:11:56 | 0:00:00 |
| Call | 1/9/2020 | 1/9/2020 11:12:02 | 1/9/2020 11:12:31 | 0:00:29 |
| Call | 1/9/2020 | 1/9/2020 13:03:07 | 1/9/2020 13:03:10 | 0:00:03 |

*Figure 5. Mobile event log.*

## 4.5 Weather and days information

As mentioned earlier, the dataset includes data for 146 days and multiple events such as weather changes, Covid pandemic and Ramadan happened during data collection. To understand the possible changes in the behaviour caused by these events, it is important to know the exact time of each event. Therefore, we have provided a file with complete information about each day such as being weekday or weekend, relation to the pandemic or Ramadan periods, weather and so on. A piece of this file is presented in Figure 6. This part of dataset can be used for cross-validation of the analysis based on dataset data and evet logs and helps to acquire more knowledge about the context.

| Date | Weekday | Weekend/Weekday | Weather | Ramadan | Covid-19 Lockdown |
|---|---|---|---|---|---|
| 1/9/2020 | Thursday | Weekend | Rain | No | No |
| 1/10/2020 | Friday | Weekend | Rain | No | No |
| 1/11/2020 | Saturday | Weekday | Cold and rain | No | No |
| 1/12/2020 | Sunday | Weekday | Cold and sunny | No | No |
| 1/13/2020 | Monday | Weekday | Cold and rain | No | No |
| 1/14/2020 | Tuesday | Weekday | Sunny | No | No |
| 1/15/2020 | Wednesday | Weekday | Cold and rain | No | No |
| 1/16/2020 | Thursday | Weekend | Cold | No | No |
| 1/17/2020 | Friday | Weekend | Cold | No | No |
| 1/18/2020 | Saturday | Weekday | Cold and rain | No | No |
| 1/19/2020 | Sunday | Weekday | Cold and rain | No | No |
| 1/20/2020 | Monday | Weekday | Cold and rain | No | No |
| 1/21/2020 | Tuesday | Weekday | Cold and rain | No | No |
| 1/22/2020 | Wednesday | Weekday | Cold and rain | No | No |
| 1/23/2020 | Thursday | Weekend | Cold and rain | No | No |

*Figure 6. Information about different various occasions and events*

## 4.6 Merged Activity and Event Log from all modalities

Considering all data sources and extracted events, this event log is created by merging and cross validating the events from all modalities. For instance, the sleep events detected by ambient sensors and wristband are cross-checked together and then using mobile event log, the awake times are identified and subtracted from the whole sleep duration. In this way, the activity and location events are created in a more accurate and multi-modal. Figure 7 shows an snapshot of this event log which includes 33848 records.

| Source | Location/Events | Date | StartTime | EndTime | Duration | Value |
|---|---|---|---|---|---|---|
| Wristband | W-Daily Steps | 1/9/2020 | 1/9/2020 0:00:00 | 1/9/2020 23:59:59 | 23:59:59 | 9939 |
| Wristband | W-Activity Time | 1/9/2020 | 1/9/2020 0:00:00 | 1/9/2020 23:59:59 | | 1:51:00 |
| Location | Kitchen | 1/9/2020 | 1/9/2020 0:50:00 | 1/9/2020 1:36:59 | 00:46:59 | |
| Location | LivingRoom | 1/9/2020 | 1/9/2020 1:37:00 | 1/9/2020 1:37:01 | 00:00:01 | |
| Location | Corridor | 1/9/2020 | 1/9/2020 1:37:02 | 1/9/2020 1:37:03 | 00:00:01 | |
| Location | Bedroom | 1/9/2020 | 1/9/2020 1:37:04 | 1/9/2020 4:44:59 | 03:07:55 | |
| RawSensor | BR-Sleeping | 1/9/2020 | 1/9/2020 1:37:04 | 1/9/2020 4:27:05 | 02:50:01 | |
| Mobile | Tools | 1/9/2020 | 1/9/2020 4:27:05 | 1/9/2020 4:27:28 | 00:00:23 | |
| Location | Corridor | 1/9/2020 | 1/9/2020 4:45:00 | 1/9/2020 4:45:01 | 00:00:01 | |
| Location | Bathroom | 1/9/2020 | 1/9/2020 4:45:02 | 1/9/2020 4:48:59 | 00:03:57 | |
| Location | Corridor | 1/9/2020 | 1/9/2020 4:49:00 | 1/9/2020 4:53:59 | 00:04:59 | |
| Location | Bedroom | 1/9/2020 | 1/9/2020 4:54:00 | 1/9/2020 8:25:59 | 03:31:59 | |
| RawSensor | BR-Praying | 1/9/2020 | 1/9/2020 4:54:03 | 1/9/2020 6:05:21 | 01:11:18 | |
| Mobile | Religious | 1/9/2020 | 1/9/2020 5:53:04 | 1/9/2020 5:53:26 | 00:00:22 | |
| RawSensor | BR-Sleeping | 1/9/2020 | 1/9/2020 6:05:21 | 1/9/2020 8:25:59 | 02:20:38 | |

*Figure 7. Multi-modal activity and event log*

## 4.7 Non-overlapping activity and location event logs

The activities, which are executed in parallel leads to additional possibilities in data mining and increased the analysis complexity. For instance, a person can use the smartphone and watch TV at the same time or use the phone during mealtime. In the non-overlapping activity and location event log, new labels generated by concatenating the labels of activity or events, which are happened in parallel, and the event log includes event records, which do not have any overlap. This event log are created in two version, one without considering the cooking activity overlaps (overlap of cooking activities are remained) (Figure 8) and the second one which removed the cooking activity overlaps as well (Figure 9).

The non-overlapping activity and location event logs with cooking overlaps contains 50368 records and the event log after removing cooking overlaps includes 50631 record.

| Source | Location/Events | Date | StartTime | EndTime | Duration | Value |
|---|---|---|---|---|---|---|
| Wristband | W-Daily Steps | 1/9/2020 | 1/9/2020 0:00:00 | 1/9/2020 23:59:59 | 23:59:59 | 9939 |
| Wristband | W-Activity Time | 1/9/2020 | 1/9/2020 0:00:00 | 1/9/2020 23:59:59 | | 1:51:00 |
| Location | Kitchen | 1/9/2020 | 1/9/2020 0:50:00 | 1/9/2020 1:36:59 | 0:46:59 | |
| Location | LivingRoom | 1/9/2020 | 1/9/2020 1:37:00 | 1/9/2020 1:37:01 | 0:00:01 | |
| Location | Corridor | 1/9/2020 | 1/9/2020 1:37:02 | 1/9/2020 1:37:03 | 0:00:01 | |
| Location | Bedroom | 1/9/2020 | 1/9/2020 1:37:04 | 1/9/2020 1:37:04 | 0:00:00 | |
| RawSensor | BR-Sleeping | 1/9/2020 | 1/9/2020 1:37:04 | 1/9/2020 4:27:05 | 2:50:01 | |
| Location | Bedroom | 1/9/2020 | 1/9/2020 4:27:05 | 1/9/2020 4:27:05 | 0:00:00 | |
| Mobile | Bedroom-Mobile | 1/9/2020 | 1/9/2020 4:27:05 | 1/9/2020 4:27:28 | 0:00:23 | Tools |
| Location | Bedroom | 1/9/2020 | 1/9/2020 4:27:28 | 1/9/2020 4:44:59 | 0:17:31 | |
| Location | Corridor | 1/9/2020 | 1/9/2020 4:45:00 | 1/9/2020 4:45:01 | 0:00:01 | |
| Location | Bathroom | 1/9/2020 | 1/9/2020 4:45:02 | 1/9/2020 4:48:59 | 0:03:57 | |
| Location | Corridor | 1/9/2020 | 1/9/2020 4:49:00 | 1/9/2020 4:53:59 | 0:04:59 | |
| Location | Bedroom | 1/9/2020 | 1/9/2020 4:54:00 | 1/9/2020 4:54:03 | 0:00:03 | |
| RawSensor | BR-Praying | 1/9/2020 | 1/9/2020 4:54:03 | 1/9/2020 5:53:04 | 0:59:01 | |
| Mobile | BR-Praying-Mobile | 1/9/2020 | 1/9/2020 5:53:04 | 1/9/2020 5:53:26 | 0:00:22 | Religious |

*Figure 8. The snapshot of non-overlapping activity and location event log with cooking overlaps*

| | | | | |
|---|---|---|---|---|
| Location | LivingRoom | 1/9/2020 | 1/9/2020 8:33:00 | 1/9/2020 8:33:01 |
| Location | Kitchen | 1/9/2020 | 1/9/2020 8:33:02 | 1/9/2020 8:33:04 |
| RawSensor | Kitchen-Cooking | 1/9/2020 | 1/9/2020 8:33:04 | 1/9/2020 8:49:12 |
| Mobile | Kitchen-Mobile-Cooking | 1/9/2020 | 1/9/2020 8:49:12 | 1/9/2020 8:51:17 |
| Location | Kitchen-Cooking | 1/9/2020 | 1/9/2020 8:51:17 | 1/9/2020 8:57:52 |
| RawSensor | K-Eating and Watching TV-Cooking | 1/9/2020 | 1/9/2020 8:57:52 | 1/9/2020 9:05:02 |
| Location | Kitchen-Cooking | 1/9/2020 | 1/9/2020 9:05:02 | 1/9/2020 9:07:59 |
| Location | LivingRoom-Cooking | 1/9/2020 | 1/9/2020 9:08:00 | 1/9/2020 9:08:01 |
| Location | Corridor-Cooking | 1/9/2020 | 1/9/2020 9:08:02 | 1/9/2020 9:08:03 |
| Location | Bathroom-Cooking | 1/9/2020 | 1/9/2020 9:08:04 | 1/9/2020 9:08:59 |

*Figure 9. The event log for all activities and locations without overlaps*

# 5  Distinctive Features of the Dataset

Compared to other datasets, our dataset offers several distinct advantages. First, it provides multimodality, integrating data from multiple devices such as ambient sensors, smartphone, and a wearable device, which allows for a more comprehensive analysis by combining information from different sources. Second, the data collection spans approximately five months, offering a longer observation period that captures significant behavioral variations over time. This extended timeframe is crucial for identifying patterns and trends that shorter datasets might miss. Additionally, our dataset covers key events known to disrupt daily routines, such as the COVID-19 pandemic with its lockdown period, New Year's holidays, and the month of Ramadan. These events introduce significant lifestyle changes, making the dataset particularly valuable for studying behavioral shifts and deviations during such times. Finally, the daily psychological reports, that are collected by a psychologist and provided beside the sensorial data, offer unique insights into the emotional and mental state of the subject, allowing researchers to explore the potential link between mood, activities, and environmental factors.

The dataset presented in this work is available for research purposes to help future advancements in smart healthcare and IoT-based systems and we invite researchers and professionals from relevant fields to use this resource in their own studies.